\documentclass[preprint,review,12pt]{elsarticle}

\usepackage{lineno}
\usepackage{graphicx}
\usepackage{amsmath}
\usepackage{amssymb}
\usepackage{setspace}
\usepackage{enumerate}
\usepackage{xcolor}
\usepackage{hyperref}
\definecolor{deepgreen}{rgb}{0,0.5,0}
\hypersetup{
    colorlinks=true,
    citecolor={deepgreen}
}
\usepackage{color}
\usepackage{array}
\usepackage[ruled,linesnumbered,vlined]{algorithm2e}
\usepackage{multirow}
\usepackage{microtype}
\usepackage[utf8]{inputenc}
\usepackage{hyperref}
\usepackage{lipsum}
\usepackage{epstopdf}
\usepackage{algorithmic}
\usepackage{subcaption}

\newcommand{\norm}[1]{\left\|#1\right\|}

\newtheorem{defn}{\noindent $\mathbf{Definition}$}[section]

\newtheorem{theorem}[defn]{$\mathbf{Theorem}$}

\newcolumntype{C}[1]{>{\centering\let\newline\\\arraybackslash\hspace{0pt}}m{#1}}

\journal{Pattern Recognition}

\begin{document}

\begin{frontmatter}

\title{Classification of the Obstructive Sleep Apnea based on X-ray images analysis by Quasi-conformal Geometry}

\author[label1]{Hei-Long Chan}
\author[label2]{Hoi-Man Yuen}
\author[label2]{Chun-Ting Au}
\author[label2]{Kate Ching-Ching Chan}
\author[label2]{Albert Martin Li}
\author[label1]{Lok-Ming Lui\corref{cor1}}

\address[label1]{Department of Mathematics, The Chinese University of Hong Kong, Hong Kong}
\address[label2]{Department of Paediatrics, Prince of Wales Hospital, The Chinese University of Hong Kong, Hong Kong}
\cortext[cor1]{Corresponding author.}
\ead{lmlui@math.cuhk.edu.hk}

\begin{abstract}
    Craniofacial profile is one of the anatomical causes of obstructive sleep apnea (OSA). By medical research, cephalometry provides information on patients' skeletal structures and soft tissues. In this work, a novel approach to cephalometric analysis using quasi-conformal geometry based local deformation information was proposed for OSA classification. Our study was a retrospective analysis based on 60 case-control pairs with accessible lateral cephalometry and polysomnography (PSG) data. By using the quasi-conformal geometry to study the local deformation around 15 landmark points, and combining the results with three linear distances between landmark points, a total of 1218 information features were obtained per subject. A $L^2$ norm based classification model was built. Under experiments, our proposed model achieves $92.5\%$ testing accuracy.
\end{abstract}


\end{frontmatter}

\section{Introduction}

Obstructive sleep apnoea (OSA) is a common sleep disorder with a reported prevalence of $35\%$ in children and is associated with cardiovascular, metabolic and neurocognitive sequelae \cite{osa1,osa2}. Craniofacial anatomy is one of the major contributing factors in OSA \cite{osa3}. Cephalometry is a relatively inexpensive, fast and readily available method that provides information on an individual’s craniofacial skeletal and soft tissue profile. Common craniofacial characteristics of OSA in children include steep mandibular plane, retrusive chin, longer lower anterior face height and smaller nasopharyngeal airway spaces \cite{osa3}. These features constitute a more restricted upper airway that poses a higher resistance and collapsibility during sleep.

The distance from the mandibular plane to hyoid bone (MP-H) is one of the most significant apnoea hypopnoea index (AHI)-correlated variables \cite{osa5,osa6,osa7,osa8}. From our previous study \cite{osa5}, MP-H significantly correlated with the presence of OSA, with an odds ratio of $2.4$ when adjusted for age, sex and BMI z-score. A significant positive correlation was also observed between MP-H and OSA severity, when comparing non-OSA group, mild and moderate-to-severe OSA groups in children. The possible relationship between lower hyoid position and OSA is theorized that a descended hyoid position is a compensatory strategy to overcome pharyngeal collapse \cite{osa9}. Another hypothesis is that the lower hyoid position is caused by enlarged tongue which contributes to airway obstruction \cite{osa10}.

Adenoid hypertrophy and posterior upper airway obstruction are intrinsic aetiology of OSA, causing narrowing of the upper airway and hence airflow restriction during sleep \cite{osa2}. They can be assessed effectively using lateral cephalogram \cite{osa11}. Adenoid size measured by adenoidal-nasopharyngeal ratio (ANR) is significantly correlated with the duration of obstructive apneas \cite{osa12} and AHI (r=0.307, p=0.034) \cite{osa6}. The minimal distance between tongue base and the nearest point on the posterior pharyngeal wall, namely the minimal posterior airway space is found to have an inverse correlation with AHI \cite{osa8}.

Traditional cephalometric analysis focuses on linear distances, angles, ratios and area of pre-identified variables \cite{osa13}. However, previous studies adopted different protocols and included
different sets of variables \cite{osa6,osa7,osa9,osa14,osa15}, although the landmarks used were mostly consistent across studies. The diagnostic value of traditional analysis remains limited that certain cephalometric predictors in paediatric population were found but have never been used as the core component for an effective OSA prediction model \cite{osa6,osa16,osa17}. Therefore, novel approach to improve cephalometric analysis is needed to enhance its diagnostic accuracy.

Quasi-conformal geometry has been proved to be an effective tool in medical analysis \cite{lui1,lui2,lui3,lui4,lui5,lui6,lui7}. In particular, it can be used in disease diagnosis such as detecting the Alzheimer's disease \cite{qc_ad,qc_ad2} by analyzing the conformality distortion on the hippocampus surface. The tool is also proved to be effective in analyzing the tooth surface for subject dating \cite{qc_tooth} as an application to bio-archaeology. We are therefore motivated to apply the quasi-conformal geometry to develop an OSA classification model.

In this work, a novel approach to cephalometric analysis for OSA classification was developed using local deformation information around manually labelled landmarks on X-ray images. Quasi-conformal geometry is useful in establishing landmark-based registration \cite{lui8,lui9,lui10}. A quasi-conformal geometry based landmark-based registration model is adopted \cite{image_reg}. The landmark- and intensity-based registration process is to find an optimal transformation between corresponding data based on specific matching features. By analysing the data at specific landmarks on the images using the quasi-conformal geometry, references of the control group and patient group can be established. For new subjects, their corresponding deformation can then be analyzed and compared against the two references. The distance of the subject's feature vector from that of the control group template and the patient group template is adopted as a classifier for disease prediction. This semi-supervised classification method aims to predict childhood OSA and potentially improve the efficacy of our current diagnostic strategies. And experiments validate that our proposed framework achieves over $92\%$ classification accuracy.

\section{Data}

\subsection{Subjects}

\begin{figure}[t]
    \centering
    \includegraphics[width=.5\textwidth]{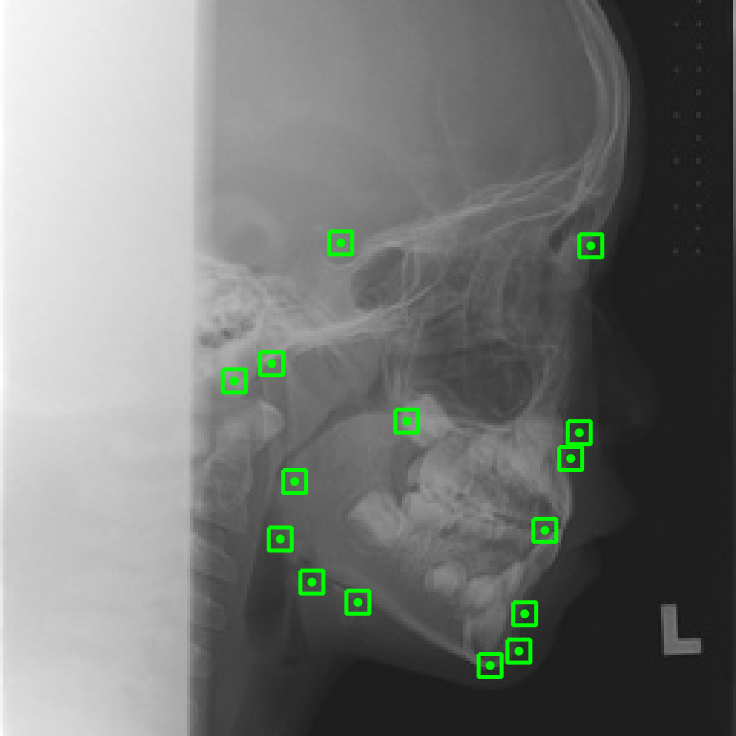}
    \caption{Demonstration of the craniofacial landmark points (green dots) and the surrounding window (green box) superimposed on a sample X-ray input image}
    \label{fig:osa_landmark}
\end{figure}

This work was a retrospective study based on 60 OSA case-control pairs who were Chinese children recruited for sleep studies in the Prince of Wales Hospital, with accessible lateral cephalometry and polysomnography (PSG) data. OSA and non-OSA groups were defined by
OAHI$\geq 1$ event/h and OAHI$<1$ event/h respectively. Lateral cephalometry was taken on the same day of admission. Patients with surgical treatment for OSA prior to cephalometry and PSG, genetic or syndromal disease,congenital or acquired neuromuscular disease, obesity secondary to an underlying cause, or craniofacial abnormalities were excluded. To study the OSA, $15$ craniofacial landmarks are labelled on each image. The landmarks are adopted from \cite{osa6} and are listed in table (\ref{tab:landmark}). Figure (\ref{fig:osa_landmark}) demonstrates the craniofacial landmarks on a reference image.

\begin{table}[h]
    \centering
    \begin{tabular}{|c|l|}
    \hline
    Landmarks & Definitions \\
    \hline
    N  & Nasion, connecting point of frontal bone and nasal bone \\
    \hline
    S  & Sella, midpoint of sella turcica\\
    \hline
    Ba & Basion, lowest point of clivus\\
    \hline
    ANS & Anterior nasal spine\\ \hline
    PNS & Posterior nasal spine\\ \hline
    A  & Deepest point of maxillary dimple\\ \hline
    B  & Deepest point of mandibular dimple\\ \hline
    \multirow{2}{*}{Gn} & Gnathion, the most anterior and inferior point on the \\ & mandibular symphysis\\ \hline
    Me & Menton, the most inferior point on the mandibular symphysis\\ \hline
    \multirow{2}{*}{Go} & Gonion, intersection of inferior margin of mandible and \\ & posterior margin of mandibular ramus\\ \hline
    \multirow{2}{*}{Ar} & Articulare, intersection of basal margin of occiput and \\ & posterior margin of mandibular ramus\\ \hline
    H  & The most anterior and superior point of hyoid bone\\ \hline
    Tant & Tip of tongue\\ \hline
    u1 & Tip of uvula\\ \hline
    Va & Vallecula\\ \hline
    \multirow{2}{*}{Phw} & The intersection of posterior pharyngeal wall and horizontal \\ & line passing hyoid bone\\ \hline
    \multirow{2}{*}{ph1} & Anterior point of a the minimal distance between tongue base \\ & and posterior pharyngeal wall\\ \hline
    \multirow{2}{*}{ph2} & Posterior point of the minimal distance between tongue base \\ & and posterior pharyngeal wall\\
    \hline
    \end{tabular}
    \caption{Definition of each craniofacial landmark adopted}
    \label{tab:landmark}
\end{table}

\subsection{Polysomnography}

The nocturnal PSG was performed at the Prince of Wales Hospital. A model SiestaTM ProFusion III PSG monitor (Compumedics Telemed, Abbotsford, Victoria, Australia) was used to record the following parameters: electroencephalogram (F4/A1, C4/A1, O2/A1), bilateral electrooculogram, electromyogram of mentalis activity and bilateral anterior tibialis. Respiratory movements of the chest and abdomen were measured by inductance plethysmography. Electrocardiogram and heart rate were continuously recorded from two anterior chest leads. Arterial oxyhaemoglobin saturation (SaO2) was measured by an oximeter with finger probe. Respiratory airflow pressure signal was measured via a nasal catheter placed at the anterior nares and connected to a pressure transducer. An oronasal thermal sensor was also used to detect the absence of airflow. Snoring was measured by a snoring microphone placed near the throat. Body position was monitored via a body position sensor.

An adequate overnight PSG is defined as recorded total sleep time of $>6$ hours. Respiratory events including obstructive apnoeas, mixed apnoeas, central apnoeas and hypopnoeas were scored based on the recommendation from the AASM Manual for the Scoring of Sleep and Associated Events [19]. Respiratory effort-related arousals (RERAs) are scored when there is a fall of $<50\%$ from baseline in the amplitude of nasal pressure signal with flattening of the nasal pressure waveform, accompanied by snoring, noisy breathing, or evidence of increased effort of breathing. A respiratory event is scored when it lasts $\geq 2$ breaths irrespective of its duration. Arousal is defined as an abrupt shift in EEG frequency during sleep, which may include theta, alpha and/or frequencies greater than 16 Hz but not spindles, with 3 to 15 seconds in duration. In REM sleep, arousal is scored only when accompanied by concurrent increases in submental EMG amplitude.

Obstructive apnoea hypopnoea index (OAHI) is defined as the total number of obstructive and mixed apnoeas and hypopnoeas per hour of sleep. Respiratory disturbance index (RDI) is defined as the total number of obstructive and mixed apnoeas, hypopnoeas and RERAs per hour of sleep. Oxygen desaturation index (ODI) is defined as the total number of dips in arterial oxygen saturation $\geq 3\%$ per hour of sleep. Arousal index (ArI) is the total number of arousals per hour of sleep. Respiratory arousal index (RAI) is the total number of arousals per hour of sleep that are associated with apnoea, hypopnoea or flow limitation. Subjects with an OAHI of $<1$/h are defined as having no OSA, while those with an OAHI between 1/h and 5/h and $\geq 5$/h are defined as having mild and moderate-to-severe OSA respectively. The PSG scoring and reporting was performed by the senior research assistant who has RPSGT qualification and experience in performing paediatric PSG. He/she was blinded to other assessment data of the subjects.

\subsection{Lateral X-ray cephalogram}
Lateral maxillofacial radiograph was taken on the same day of admission to overnight PSG. All radiographic examination was performed with Direct Digital Radiography System (Carestream DRX-1 Evolution DR System, US) using standardized protocol (70-75kVp, Automatic sensor of around 6-10 mAs, 150-cm film-focus distance).

\section{Mathematical Background}

In this section, the quasi-conformal theory is reviewed as it is the key concept towards our proposed model. It is the foundation of our registration model and it also contributes to our proposed feature, the conformality distortion, to classify OSA.

\subsection{Review on quasi-conformal geometry on 2D domain}
\label{subsec:ReviewQCGeometry}

Let $\Omega_1$ and $\Omega_2$ be two rectangular image domain, which are regarded as subsets of $\mathbb{C}$.  A diffeomorphism $f:\Omega_1\rightarrow\Omega_2$ is defined to be conformal if it is a complex function satisfying the Cauchy-Riemann equation
\begin{equation}
\frac{\partial f}{\partial \bar{z}}=0,
\end{equation}
where $\frac{\partial}{\partial \bar{z}}=\frac{\partial}{\partial x}+i\frac{\partial}{\partial y}$. A conformal mapping always preserves angles and thus the local geometry is preserved under the mapping.

Then, an orientation-preserving homeomorphism $f:\Omega_1\rightarrow\Omega_2$ is defined to be quasi-conformal if it satisfies the Beltrami equation
\begin{equation}
\frac{\partial f(z) }{\partial \bar{z}}=\mu(z)\frac{\partial f(z)}{\partial z},
\label{BE}
\end{equation}
where $\mu:\Omega_1\rightarrow \mathbb{C}$ is Lebesgue measurable satisfying $||\mu ||_{\infty}<1$, and $\frac{\partial}{\partial z}=\frac{\partial}{\partial x}-i\frac{\partial}{\partial y}$.

\begin{figure}[t]
  \centering
  \includegraphics[width=0.4\textwidth]{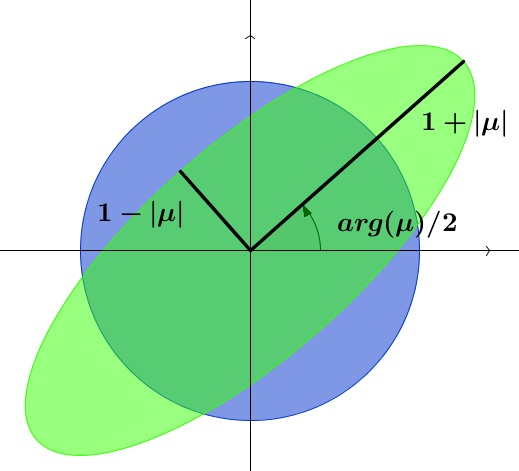}
  \caption{Illustration of the conformality distortion in 2-dimensional space: mapping a infinitesimal disk (blue) to an infinitesimal ellipse (green). The disk and the ellipse are rescaled for illustrative purpose}
  \label{cdA}
\end{figure}

Obviously, $||\mu ||_{\infty}=0$ if and only if $f$ is conformal. Hence, the notion of quasi-conformal maps is a generalization of conformal maps. Infinitesimally, suppose $0\in\Omega_1$, then for any $z\in Nbd(0,\delta)$ where $\delta>0$ is small, a quasi-conformal mapping $f$ has the following local parametric expression
\begin{equation}
\begin{split}
f(z) & \approx f(0)+f_{z}(0)z+f_{\bar{z}}(0)\bar{z} \\
     & =f(0)+f_{z}(0)(z+\mu (0)\bar{z}). \\
\end{split}
\end{equation}
Note that the translation function $f(0)$ and the dilation function $f_{z}(0)$ are conformal, so all the non-conformality of $f$ is due to the term $D(z)=z+\mu (0)\bar{z}$ which causes $f$ to map an infinitesimal circle to an infinitesimal ellipse (See Figure (\ref{cdA})).

Hence, the study of non-conformality reduces to the analysis of the term $\mu$, which is called the {\it Beltrami coefficient}. In fact, for any $p\in\Omega$, the angle of maximal magnification is $arg(\mu (p))/2$ with magnifying factor $1+|\mu (p)|$ while the angle of maximal contraction is the orthogonal angle $(arg(\mu (p))-\pi )/2$ with contraction factor $1-|\mu (p)|$.

Indeed, by defining $\mu_f$ for a complex function $f$ using the equation (\ref{BE}), it can be seen that $\mu_f$ can be used to distinguish orientation preserving homeomorphisms.
\begin{theorem}
Let $f:\mathbb{C}\to\mathbb{C}$ be a complex mapping. Define
\begin{equation}
\mu_f=\frac{\partial f}{\partial\bar{z}}\bigg /\frac{\partial f}{\partial z},
\end{equation}
then $\norm{\mu_f}_{\infty}<1$ if and only if $f$ is an orientation preserving homeomorphism.
\end{theorem}
Here, $\mu_f(x)$ is called the conformality distortion of the function $f$ at $x$. Its magnitude and angle can be used to determine the ``distance'' of $f$ from being conformal.

Note that there is a one-to-one correspondence between a quasi-conformal mapping $f$ and its Beltrami coefficient $\mu$. Given $f$, there exists a Beltrami coefficient $\mu$ such that $(f,\mu)$ satisfies the Beltrami equation. Conversely, the following theorem states that given an admissible Beltrami coefficient $\mu$, there always exists an quasi-conformal mapping $f$ associating to this $\mu$.

\begin{theorem}[Measurable Riemannian Mapping Theorem]
Suppose $\mu:\mathbb{C}\to\mathbb{C}$ is Lebesgue measurable satisfying $\norm{\mu}_{\infty}<1$, then there exists a quasi-conformal homeomorphism $f$ from the unit disk to itself, which is in the Sobolev space $W^{1,2}(\mathbb{C})$ and satisfies the Beltrami equation in the distribution sense. Furthermore, assuming the mapping is stationary at $0$, $1$ and $\infty$, the associated quasi-conformal homeomorphism $f$ is uniquely determined.
\end{theorem}

Therefore, under suitable normalization, a homeomorphism from $\mathbb{C}$ to $\mathbb{C}$ can be uniquely determined by its associated Beltrami coefficient. This is a crucial property that allows one to register between two images by homeomorphisms, which can be constructed by applying constraints on the Beltrami coefficient corresponding to the registration mapping.

\section{Proposed Model}

In this work, we propose to analyze the deformation between X-ray images of skulls to detect OSA. In the first subsection, we discuss the image registration with reference to the craniofacial landmarks. Then, geometric distortions of the deformation are calculated to form a feature vector for each subject, which is the main content of the second subsection. Finally, we develop a classification model using the discriminating feature vectors.

\subsection{Image Registration}

A landmark-matching registration model is adopted for computing the mutual correspondence between subjects \cite{image_reg}. More specifically, the model develops the registration mapping between images $I_i,I_j:\Omega\to\mathbb{R}$ of subjects $i,j$ by minimizing the energy
\begin{equation}
    E(\mu,f)=\int_{\Omega}|\nabla\mu|^2+\alpha\int_{\Omega}|\mu|^2+\beta\int_{\Omega}(I_i-I_j\circ f)^2.
\label{osa_im_reg}
\end{equation}
The registration mapping is a smooth homeomorphism matching the intensity between $I_i,I_j$. To incorporate with the landmark constraints, a splitting variables scheme is used and the corresponding minimization problem is
\begin{equation}
    E(\mu,\nu,f)=\int_{\Omega}|\nabla\nu|^2+\alpha\int_{\Omega}|\nu|^2+\sigma\int_{\Omega}|\nu-\mu|^2+\beta\int_{\Omega}(I_i-I_j\circ f^\mu)^2,
\label{osa_im_reg_split}
\end{equation}
in which $\mu$ is the Beltrami coefficient of $f^\mu$ and $\mu$ is forced to match with $\nu$ by the third term in (\ref{osa_im_reg_split}).Using the formulation (\ref{osa_im_reg_split}), the landmark constraints can be added to the variational model by constructing a Beltrami coefficient corresponding to a mapping $g$, which aligns the landmarks exactly and closely resembles $\mu$, in the alternating minimization of $(\ref{osa_im_reg_split})$. This process is done by the Linear Beltrami Solver (LBS). Fore more details about the formulation of the registration model, readers are referred to \cite{image_reg}. The application of the quasi-conformal registration is beneficial to reduce the calibration error in taking the X-ray photos for each subject. In other words, the effect of global scaling, global rotation and global linear translation are minimized by quasi-conformal mappings.

\subsection{Classification Features}

Now, suppose we have $N$ subjects in the database, in which the first $N/2$ subjects are in the control class and the last $N/2$ subjects are in the OSA class. As for disease classification, it is common to construct a template subject for the control class. In this work, we propose to construct such template in the space of Beltrami coefficients. In particular, we randomly pick a control subject $I=I_i$ as the reference subject. Each of the images in the database is registered to $I$ by the above registration model. Let $f_{i}$ be the registration mapping aligning each craniofacial landmark vertex $v_k$ on $I$ to the corresponding vertex $v_{ki}$ on the subject $i$. For each landmark point $v_k$ on the template object, a square window centered at $v_k$ of size $w$ is extracted. Thus, each of the $15$ windows includes $w^2$ vertex points. Figure (\ref{fig:osa_landmark}) demonstrates the windows at each landmark point. At each vertex point $v$ included, the magnitude $|\mu(v)|$ and the argument $\text{arg}(\mu(v))$ of the Beltrami coefficient $\mu$ of $f_i$ is computed. We construct the template deformation to be the mean of the Beltrami coefficient among the control class, that is,
\begin{equation*}
    \mu_{template}(v) = \frac{\sum^{N/2}_{i=1}\mu(v_i)}{N/2}.
\end{equation*}

To construct features for the classification model, we linearly combine $|\mu|$ and $\text{arg}(\mu)$ at each vertex to describe the local deformation around the vertex. That is, we define the deformation index:
\begin{equation}
    E^i_{deform}(v)=\alpha\cdot|\mu_i(v)|+\beta\cdot\frac{\text{arg}(\mu_i(v))}{\pi}
\label{osa_deform_index}
\end{equation}
for the subject $i$, where $\alpha,\beta>0$. Note that since $|\mu(v)|\in [0,1]$ and $\text{arg}(\mu(v))\in [0,\pi]$ for every vertex, so a normalization by $1/\pi$ is added to the latter term to balance the contribution of the two measurements towards $E_{deform}$. The parameters $\alpha,\beta$ are chosen such that $\alpha^2+\beta^2=1$. The detail of this part will be elaborated in a latter session.

It is noted that among those craniofacial landmarks, some of them (i.e. Phw, ph1, and ph2) which are on the pharyngeal wall cannot be compared directly among subjects. In this work, we incorporate the mutual distances between each pair of them as features for the classification. That is, we include the distance $d^1_i$ from the mandibular plane to the hyoid bone (MP-H), the distance $d^2_i$ from the hyoid bone to the posterior pharyngeal wall (H-Phw) and the lower pharyngeal width $d^3_i$ (ph1-ph2) in the deformation index.

Incorporating the $3$ distance measurements with the deformation index $E_{deform}$, each subject $i$ is now described by the feature vector
\begin{equation}
    C_i=[E^i_{deform}(v_1),E^i_{deform}(v_2),\dots,E^i_{deform}(v_{15\times w^2}),\bar{d}^1_i,\bar{d}^2_i,\bar{d}^3_i],
\label{osa_featurevector}
\end{equation}
where $\bar{d}^j_i$ is the normalization of $d^j_i$ across subjects such that $\text{max}(\bar{d}^j_i)=1$ for all $i$, for each $j=1,2,3$. In this work, we choose the window size to be $w=9$, so each subject is represented by $15\cdot 9^2=1215$ deformation index, together with $3$ distance measurements. It is noted that if the windows at two different landmark vertices on the same subject overlap with each other, some vertices in the windows will have multiple contribution to $C_i$.

To further improve the discriminating power of the feature vector, a t-test incorporating the bagging predictors \cite{bagging} is applied to trim the feature vector (\ref{osa_featurevector}) with respect to the deformation index. In the traditional t-test, a probability $p_k$ called the p-value is defined and computed for each feature $E_{deform}(v_k)$ which evaluates the discriminating power of the corresponding feature in separating the given two classes. The bagging predictors strategy further improve the stability of the t-test by a leave-one-out scheme. More specifically, a total of $N$ testes are performed. In each test, the $i$-th subject is excluded temporarily and the t-test is applied on the remaining subjects. This gives the p-value $p^i_k$ for the feature $k$ in the $i$-th iteration. After all the $N$ testes, the p-value of a feature is computed by
\begin{equation}
    p_k=\min_ip^i_k.
\end{equation}
Finally, the features with low discriminating power can be expelled from our classification machine by choosing only the $K$ features with high discriminating power in order.

\begin{figure}[t]
    \centering
    \includegraphics[width=\textwidth]{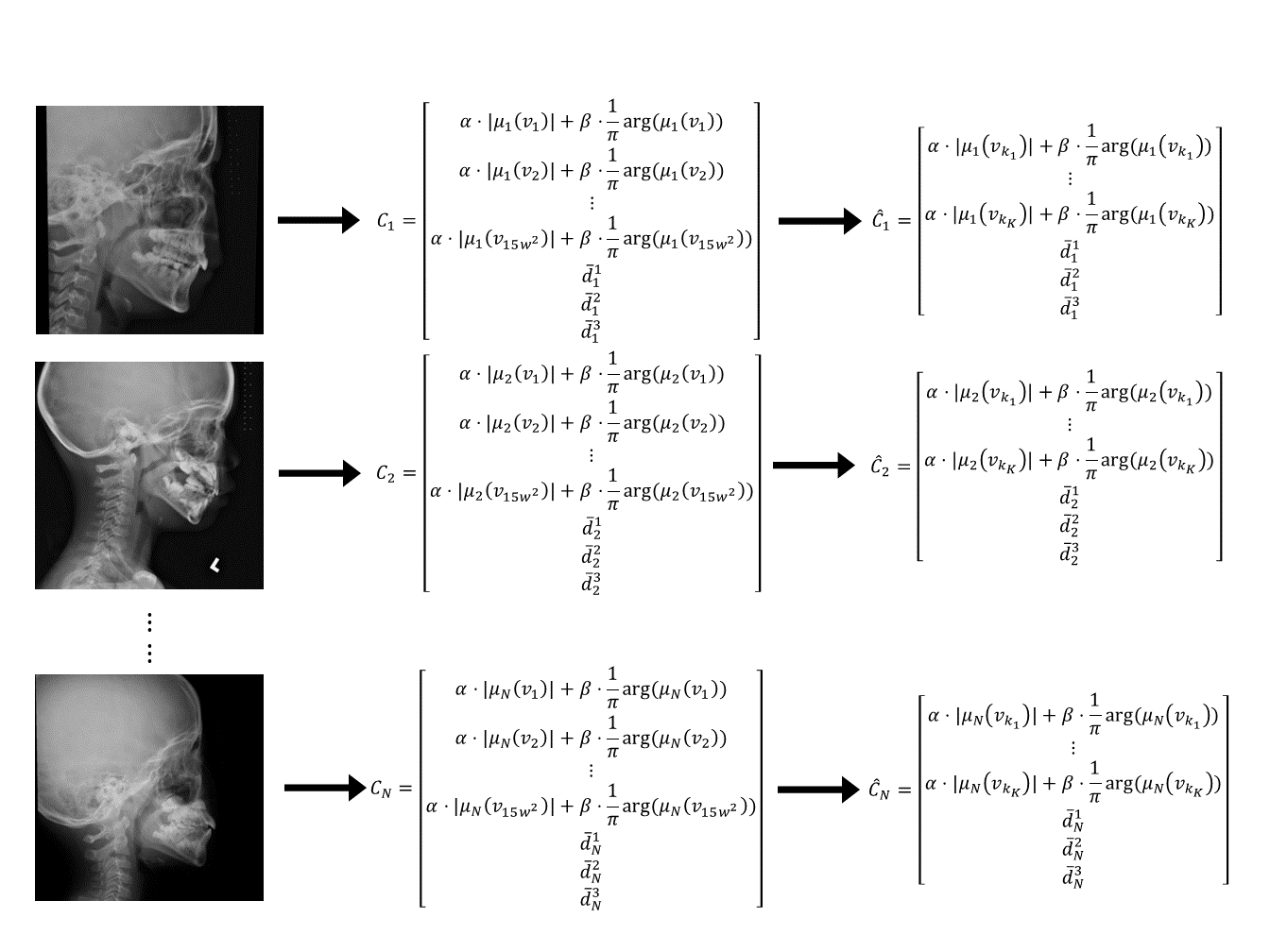}
    \caption{Illustration of the process generating the discriminating feature vector for each subject}
    \label{fig:osa_feature_vector_pipeline}
\end{figure}

Therefore, our model uses the discriminating feature vector
\begin{equation}
    \hat{C}_i=[E_{deform}(v_{k_1}),E_{deform}(v_{k_2}),\dots,E_{deform}(v_{k_K}),\bar{d}^1_i,\bar{d}^2_i,\bar{d}^3_i]
\label{osa_feature_vector_trim}
\end{equation}
as the input to our classification machine. Figure (\ref{fig:osa_feature_vector_pipeline}) illustrates the pipeline generating the discriminating feature vector for each subject from the corresponding deformation mapping to the template subject.

\subsection{Classification Machine}

Now, we can build the classification model. In this work, we propose to apply a simple $L^2$-norm based classification model which is also used in \cite{qc_ad}. We first compute the mean of the feature vectors among the NC class:
\begin{equation}
    C_{mean} = \text{mean}(\hat{C}_1,\hat{C}_2,\dots,\hat{C}_{N/2}).
\end{equation}
Then, the $L^2$ distance between the feature vector of each subject $i=1,\dots,N$ and the mean feature vector $C_{mean}$ is computed:
\begin{equation}
    d_i = ||\hat{C}_i-C_{mean}||_2.
\end{equation}
Since we assume that subjects from the control class should possess similar geometry of the skull, the deformation from a control subject $i$ to the chosen template subject $I$ should be small. That is, $d_i$ should be small if $i\leq N/2$. By sorting $\{d_1,d_2,\dots,d_N\}$, there exists an optimal cutting threshold $d_{opt}>0$ maximizing the number of members in the set
\begin{equation}
    \{i\in[1,\frac{N}{2}]:d_i<d_{opt}\}\cup\{i\in[\frac{N}{2}+1,N]:d_i>d_{opt}\}.
\end{equation}
That is, $d_{opt}$ is the optimal threshold separating the control class and the OSA class. This gives a classification machine providing an automatic diagnosis for a new subject.

Suppose a new subject is given, to predict if it belongs to the control class or the OSA class, we compute its corresponding feature vector $\hat{C}_{new}$ as in (\ref{osa_feature_vector_trim}) and hence the distance
\begin{equation}
    d_{new}=||\hat{C}_{new}-C_{mean}||_2.
\end{equation}
Then, if $d_{new}<d_{opt}$, we conclude the subject belongs to the control class. Otherwise if $d_{new}>d_{opt}$, we conclude the subjects belongs to the OSA class.

\subsection{Parameter Optimization}

The parameters $\alpha,\beta$ in the deformation index (\ref{osa_deform_index}) can be automatically optimized to maximize the accuracy of the model. It is based on the fact that the discriminating power of the deformation index $E_{deform}$ is invariant under normalization. Therefore, we can constraint the parameter space to lie within the unit circle. In other words, we search for the optimal $(\alpha_{opt},\beta_{opt})$ in the space
\begin{equation}
    \{(\alpha,\beta)\in\mathbb{R}^2:\alpha>0,\beta>0,\alpha^2+\beta^2=1\}.
\end{equation}
Using the spherical coordinates, we can set a density parameter $\rho\in(0,\frac{1}{2}]$ and compute
\begin{equation}
    \alpha_k=\cos{k\rho\pi},\quad\beta_k=\sin{k\rho\pi},\quad k=0,1,\dots,\frac{1}{2\rho}.
\end{equation}
Each pair of $(\alpha_k,\beta_k)$ varies the contribution of $|\mu|$ and $\text{arg}(\mu)$ to the deformation index $E_{deform}$ and hence gives a different classification model. The accuracy of each model can then be tested by the $10$-fold cross validation and thus the optimal parameter $(\alpha_{opt},\beta_{opt})$ can be chosen to be the one contributing to the model of the highest validation accuracy. It is noted that the number $K$ in choosing the discriminating features has to be optimized by hand-tuning.

\section{Experiments results}

In this work, we are given $120$ subjects consisting of $60$ control subjects and $60$ OSA subjects. To test the accuracy of the proposed model, we perform $100$ testes. In each test, we randomly pick $40$ control subjects and $40$ OSA subjects to compose a sub-database to train the classification model. That is, we apply the $10$-fold cross validation onto the sub-database (of size $40$) to optimize the parameters $(\alpha,\beta)$. In a $10$-fold cross validation, the database is partitioned into $10$ equal portions and $10$ sub-experiments are performed. In each sub-experiment, one portion is excluded and the classification model is built using the remaining data. Afterwards, the subjects in the excluded portion is used to serve as testing subjects. In this manner, each data in the database serve as a testing subject for exactly once and an overall classification accuracy of all the $10$ sub-experiments can be calculated. $10$-fold cross validation is a very popular validation method to evaluate the accuracy of a classification model if only a small database is given.

For each of the $100$ testes, a $10$-fold cross validation is performed on the sub-database and the optimal parameters $(\alpha_{opt},\beta_{opt})$ are obtained. Then, the classification machine is tested with the remaining $20$ control subjects and the $20$ OSA subjects. This gives a testing accuracy of the proposed machine. The results of the $100$ testes are combined to evaluate the mean accuracy of the proposed OSA classification machine.

The highest classification accuracy is $92.5\%$ (sensitivity $95\%$ and specificity $90\%$) achieved at choosing $K=500$. That is, $500$ vertices out of the $1,218$ vertices having the highest discriminating power is chosen. Table (\ref{tab:osa_accuracy}) records the classification accuracy of the proposed model in choosing different $K$.

\begin{table}[h]
    \centering
    \begin{tabular}{|c|c|c|c|c|}
        \hline
        No. of features & $(\alpha_{opt},\beta_{opt})$ & Sensitivity & Specificity & Accuracy \\
        \hline
        $500$ & $(0.985,0.173)$ & $95.0\%$ & $90.0\%$ & $92.5\%$ \\
        \hline
        $800$ & $(0.996,0.089)$ & $89.3\%$ & $85.7\%$ & $87.5\%$ \\
        \hline
        $1215$ & $(0.989,0.150)$ & $72.6\%$ & $62.6\%$ & $67.6\%$ \\
        \hline
    \end{tabular}
    \caption{Statistics of the classification accuracy of the proposed model}
    \label{tab:osa_accuracy}
\end{table}

\begin{figure}[t]
    \centering
    \includegraphics[width=.9\textwidth]{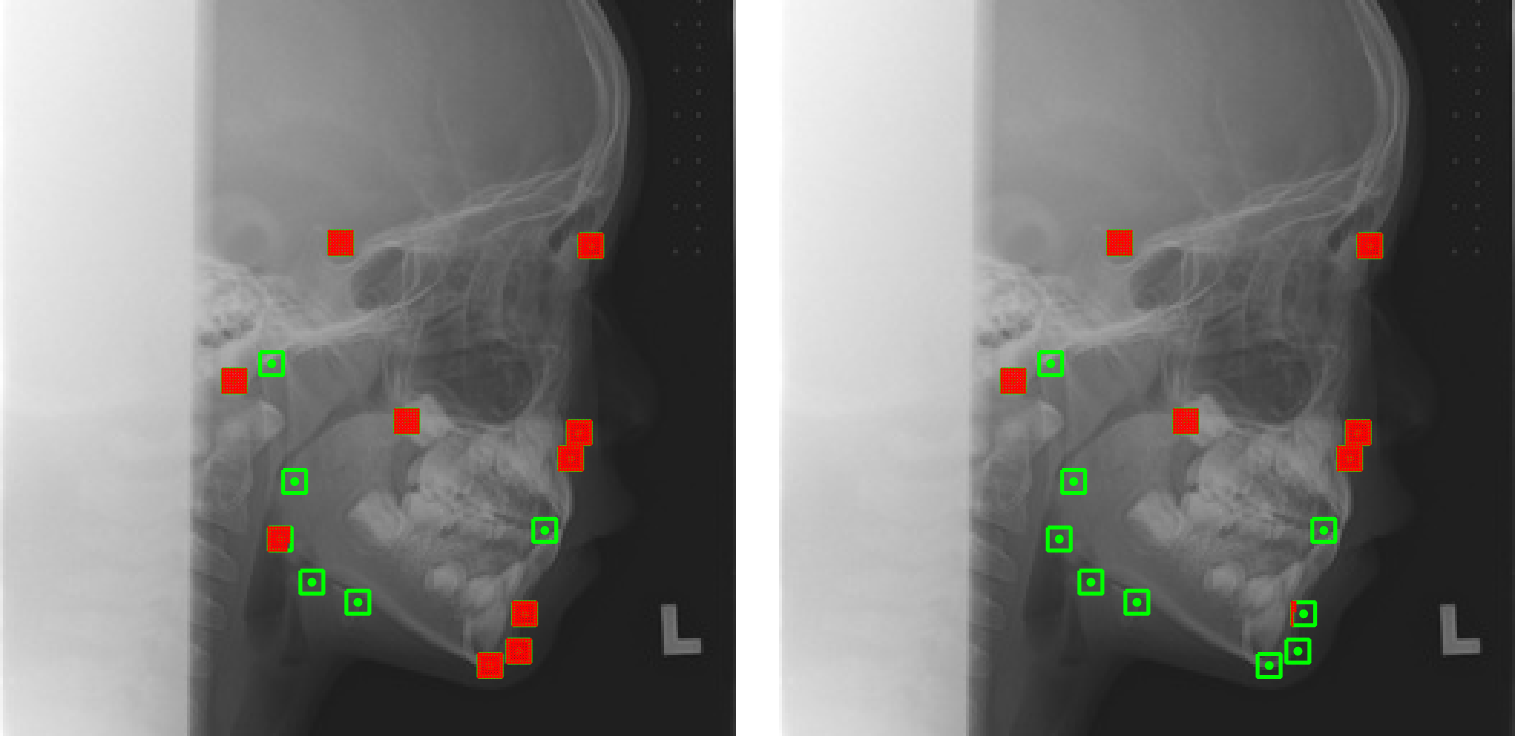}
    \caption{Visualization of the vertices picked (red) by the model in constructing the classification model. (Left) $K=800$; (Right) $K=500$.}
    \label{fig:feature_visualize}
\end{figure}

According to the automatic optimization of the coefficients $(\alpha,\beta)$, the magnitude $|\mu|$ of the Beltrami coefficient $\mu$ of the deformation has a consistently higher discriminating power over the argument $\arg(\mu)$ of $\mu$ in the classification model. This can be explained by the fact that the magnitude $|\mu|$ describes the degree of non-conformal distortion while the argument $\text{arg}(\mu)$ describes the direction of non-conformal distortion.

Figure (\ref{fig:feature_visualize}) plots the $K$ vertices with the highest discriminating power.  From the figure, it can be seen that some craniofacial landmarks do have a higher discriminating power to the classification model. This is also a contribution of our model to help validating the discriminating power of each craniofacial landmark in the OSA diagnosis.

\subsection{Comparison with other methods}

In literature of OSA studies using lateral cephalometry,  the majority was comparing the linear distance, angles and ratios measured directly on the cephalogram between the OSA group and control group. To compare our proposed model with the conventional methods, we built another OSA classification machine using the same database.

Twenty-two cephalometric parameters were measured based on the landmarks (listed in table (\ref{tab:measurement}) and table (\ref{tab:measurement2})). The measurements are stacked to form the feature vector for each subject. Then, we apply the SVM to create the classification model. For fair testing, the 10-fold cross validation is used to test the model with 60 control-OSA pairs of subjects randomly selected from the database. And a total of $100$ testes are performed to neutralize possible bias to a certain data separation.

\begin{table}[h]
    \centering
    {\renewcommand\arraystretch{1.25}
    \begin{tabular}{|c|c|c|}
    \hline
    Categories & Measurements & Definitions \\
    \hline
    \multirow{10}{3cm}{Nasal cavity and nasopharyngea 1 space}  & \multirow{2}{*}{Ba-N} & \multicolumn{1}{p{6cm}|}{\raggedright The distance from the lowest point of clivus to nasion}\\
    \cline{2-3}
    & S-N & \multicolumn{1}{p{6cm}|}{\raggedright The distance from sella to nasion}\\
    \cline{2-3}
    & \multirow{2}{*}{Ba-S} & \multicolumn{1}{p{6cm}|}{\raggedright The distance from the lowest point of clivus to sella}\\
    \cline{2-3}
    & \multirow{2}{*}{Ba-S-N} & \multicolumn{1}{p{6cm}|}{\raggedright The angle between the lowest point of clivus, sella, and nasion}\\
    \cline{2-3}
    & \multirow{3}{*}{Ba-S-PNS} & \multicolumn{1}{p{6cm}|}{\raggedright The angle between the lowest point of clivus, sella, and posterior nasal spine}\\
    \hline
    \multirow{11}{3cm}{Position of hyoid bone} & \multirow{2}{*}{MP-H} & \multicolumn{1}{p{6cm}|}{\raggedright The distance from mandibular plane to hyoid bone}\\
    \cline{2-3}
    & \multirow{2}{*}{Gn-Go-H} & \multicolumn{1}{p{6cm}|}{\raggedright The angle betwenn the line Gn-Go and the line Go-H}\\
    \cline{2-3}
    & \multirow{5}{*}{MP-H/Go-Gn} & \multicolumn{1}{p{6cm}|}{\raggedright The position of hyoid bone, the ratio of the distance between mandibular plane and hyoid bone and the length of mandibular body}\\
    \cline{2-3}
    & \multirow{2}{*}{H-Phw} & \multicolumn{1}{p{6cm}|}{\raggedright  The distance between hyoid bone and posterior pharyngeal wall}\\
    \hline
    \multirow{5}{3cm}{Soft tissue} & ul-PNS & \multicolumn{1}{p{6cm}|}{\raggedright The length of soft palate} \\
    \cline{2-3}
    & Va-Tant & \multicolumn{1}{p{6cm}|}{\raggedright The length of tongue}\\
    \cline{2-3}
    & \multirow{3}{*}{ph1-ph2} & \multicolumn{1}{p{6cm}|}{\raggedright  The minimal distance between tongue base and posterior pharyngeal wall}\\
    \hline
    \end{tabular}}
    \caption{List of all cephalometric measurements adopted to the conventional classification machine (part A)}
    \label{tab:measurement}
\end{table}

\begin{table}[t]
    \centering
    {\renewcommand\arraystretch{1.25}
    \begin{tabular}{|c|c|c|}
    \hline
    Categories & Measurements & Definitions \\
    \hline
    \multirow{24}{3cm}{Maxilla and mandible} & Go-Gn & \multicolumn{1}{p{6cm}|}{\raggedright The length of mandibular body}\\
    \cline{2-3}
    & \multirow{3}{*}{MP} & \multicolumn{1}{p{6cm}|}{\raggedright Mandibular plane, tangent to the lower border of the mandible through menton}\\
    \cline{2-3}
    & \multirow{2}{*}{SN-GoGn} & \multicolumn{1}{p{6cm}|}{\raggedright The angle between S-N line and Go-Gn line}\\
    \cline{2-3}
    & PNSANS-GoGn & \multicolumn{1}{p{6cm}|}{\raggedright The angle between maxilla and mandible}\\
    \cline{2-3}
    & \multirow{3}{*}{S-N-A} & \multicolumn{1}{p{6cm}|}{\raggedright  The angle between sella, nasion, and deepest point of maxillary dimple}\\
    \cline{2-3}
    & \multirow{3}{*}{S-N-B} & \multicolumn{1}{p{6cm}|}{\raggedright The angle between sella, nasion, and deepest point of mandibular dimple}\\
    \cline{2-3}
    & \multirow{4}{*}{A-N-B} & \multicolumn{1}{p{6cm}|}{\raggedright he angle between the deepest point of maxillary dimple, nasion, and deepest point of mandibular dimple}\\
    \cline{2-3}
    & \multirow{2}{*}{Ar-Go-Gn} & \multicolumn{1}{p{6cm}|}{\raggedright The angle between the line Ar-Go and the line Go-Gn}\\
    \cline{2-3}
    & \multirow{2}{*}{Ar-Go-N} & \multicolumn{1}{p{6cm}|}{\raggedright The angle between the line Ar-Go and the line Go-N} \\
    \cline{2-3}
    & \multirow{2}{*}{N-Go-Gn} & \multicolumn{1}{p{6cm}|}{\raggedright The angle between the line N-Go and the line Go-Gn}\\
    \hline
    \end{tabular}}
    \caption{List of all cephalometric measurements adopted to the conventional classification machine (part B)}
    \label{tab:measurement2}
\end{table}

The accuracy of the model using conventional cephalometric parameters is $70.3\%$. If the top-10 best features among the parameters are extracted (using the bagging-incorporated t-test strategy as in our proposed model), the accuracy is $74.6\%$. Comparing the accuracy, it is evident that our proposed QC-based model really contributes to a more accurate classification of OSA. This can be explained by the fact that the conformality distortion provides a deeper infinitesimal understanding of the underlying deformation than conventional cephalometric measurements.

\section{Conclusion}

A new approach to cephalometric analysis using quasi-conformal geometry based local deformation information is proposed for classifying the Obstructive Sleep Apnea (OSA). The proposed model combines information from the conformality distortion with the distance measurements between several craniofacial landmarks to formulate a feature vector to describe each subject. A t-test incorporating the bagging predictor is applied to trim the feature vector and increase its discriminating power. A $L^2$-norm based classification machine is built using the trimmed feature vector. Under experiments on a database consisting of $60$ OSA case-control pairs, our proposed model achieves $92.5\%$ accuracy in choosing the top $500$ best features. In the future, we will apply the current framework in the neural network setting to further improve the accuracy and efficiency.

\vspace{2mm}
\begin{scriptsize}

\end{scriptsize}

\end{document}